# Multimodal Limbless Crawling Soft Robot with a Kirigami Skin

Jonathan Tirado[1], Aida Parvaresh[1], Burcu Seyidoğlu[1], Darryl A. Bedford[2], Jonas Jørgensen[1], Ahmad Rafsanjani[1,*]

[1] SDU Soft Robotics, Biorobotics Section, The Maersk McKinney Moller Institute, University of Southern Denmark, Odense 5230, Denmark

[2] Drawstring Origami Ltd., London, UK

* Corresponding author ahra@sdu.dk

Date: May 5, 2025

**Abstract**

Limbless creatures can crawl on flat surfaces by deforming their bodies and interacting with asperities on the ground, offering a biological blueprint for designing efficient limbless robots. Inspired by this natural locomotion, we present a soft robot capable of navigating complex terrains using a combination of rectilinear motion and asymmetric steering gaits. The robot is made of a pair of antagonistic inflatable soft actuators covered with a flexible kirigami skin with asymmetric frictional properties. The robot's rectilinear locomotion is achieved through cyclic inflation of internal chambers with precise phase shifts, enabling forward progression. Steering is accomplished using an asymmetric gait, allowing for both in-place rotation and wide turns. To validate its mobility in obstacle-rich environments, we tested the robot in an arena with coarse substrates and multiple obstacles. Real-time feedback from onboard proximity sensors, integrated with a human-machine interface (HMI), allowed adaptive control to avoid collisions. This study highlights the potential of bioinspired soft robots for applications in confined or unstructured environments, such as search-and-rescue operations, environmental monitoring, and industrial inspections.

**Introduction**

Earthworms can crawl on the ground and burrow underground without the use of limbs. For limbless locomotion on flat surfaces, the absence of push points over the surface requires the coordination of body deformation and static friction to generate propulsive forces. The rhythmic contraction of earthworms' muscles produces peristaltic waves along their slender bodies [1] while friction-enhancing bristles on their skin, called setae, ensure a firm grip on the ground with each stride [2, 3]. The setae generate a directionally asymmetric friction that is easy to overcome in the direction of movement but strong enough to prevent sliding back. Thus, three fundamental elements of limbless locomotion on terrains with uniform roughness are large deformability, rhythmic contractions, and asymmetric friction. The limbless locomotion of earthworms has inspired the development of several crawling soft robots that replicate some of their morphological features, enabling them to crawl on uniform terrains [4, 5, 6], inside pipes [7, 8, 9], and through granular media [10, 11]. However, unifying all of these in a crawling robot remains unexplored. Additionally, many earthworm-inspired soft robots can only move in a straight line and do not possess steering capabilities, which limits their applicability to unstructured real-world terrains.

To replicate body deformation, several researchers have developed worm-inspired soft robots powered by various actuation mechanisms. For example, Soek *et al*. developed a peristaltic soft robot powered by shape memory alloy nickel titanium coils integrated into an elastic tubular braided mesh to generate antagonistic axial and radial contractions for rectilinear locomotion on flat surfaces [12]. Das *et al*. developed a modular soft robot for locomotion in multi-terrain environments using a fluid-driven peristaltic soft actuator capable of two active configurations by alternating between positive and negative pressure inputs, producing longitudinal forces for axial penetration and radial forces for anchorage through bidirectional deformation of its central bellows-like structure [13]. Yoon *et al*. created an untethered soft earthworm robot that operates based on the thermally controlled gas-liquid phase transition of a soft thermoelectric pneumatic device [14]. Calderón *et al*. developed a pneumatic soft robotic system consisting of two expanding ends connected to an extending actuator, replicating the motion and functionality of a single burrowing earthworm segment, enabling locomotion inside horizontal, inclined, and vertical pipes [15]. Zhang *et al*. further advanced this design by incorporating three chambers in the central actuator for navigation and imaging through multi-branch tubular phantoms [16].

To incorporate asymmetric friction, researchers integrated passive hooks and bristles [13, 17, 18], multi-material casted setae [19], and 3d printed scales [20] into soft actuators, creating low friction in the movement direction and high friction in the backward direction, thereby enabling two-anchor crawling [21]. Kirigami is a traditional papercraft technique that enables the transformation of thin, flat sheets into complex three-dimensional structures through patterned cuts, offering functional and flexible skins for crawling robots. An effective strategy for introducing friction asymmetry in soft crawling robots involves covering soft actuators with kirigami skins created by introducing asymmetric repetitive cut patterns into plastic films, where the elongation of the underlying actuator triggers directionally aligned pop-ups [22]. These kirigami skins have enabled bioinspired soft robots to achieve rectilinear locomotion [23, 24], burrow through cohesive soil [25] and move by lateral undulation [26]. However, while such stretchable kirigami skins coordinate body deformation with friction asymmetry, existing designs are often limited to uniaxial movement because bending-induced compressive forces during robot steering result in nonuniform and unpredictable deformation of the kirigami skins.

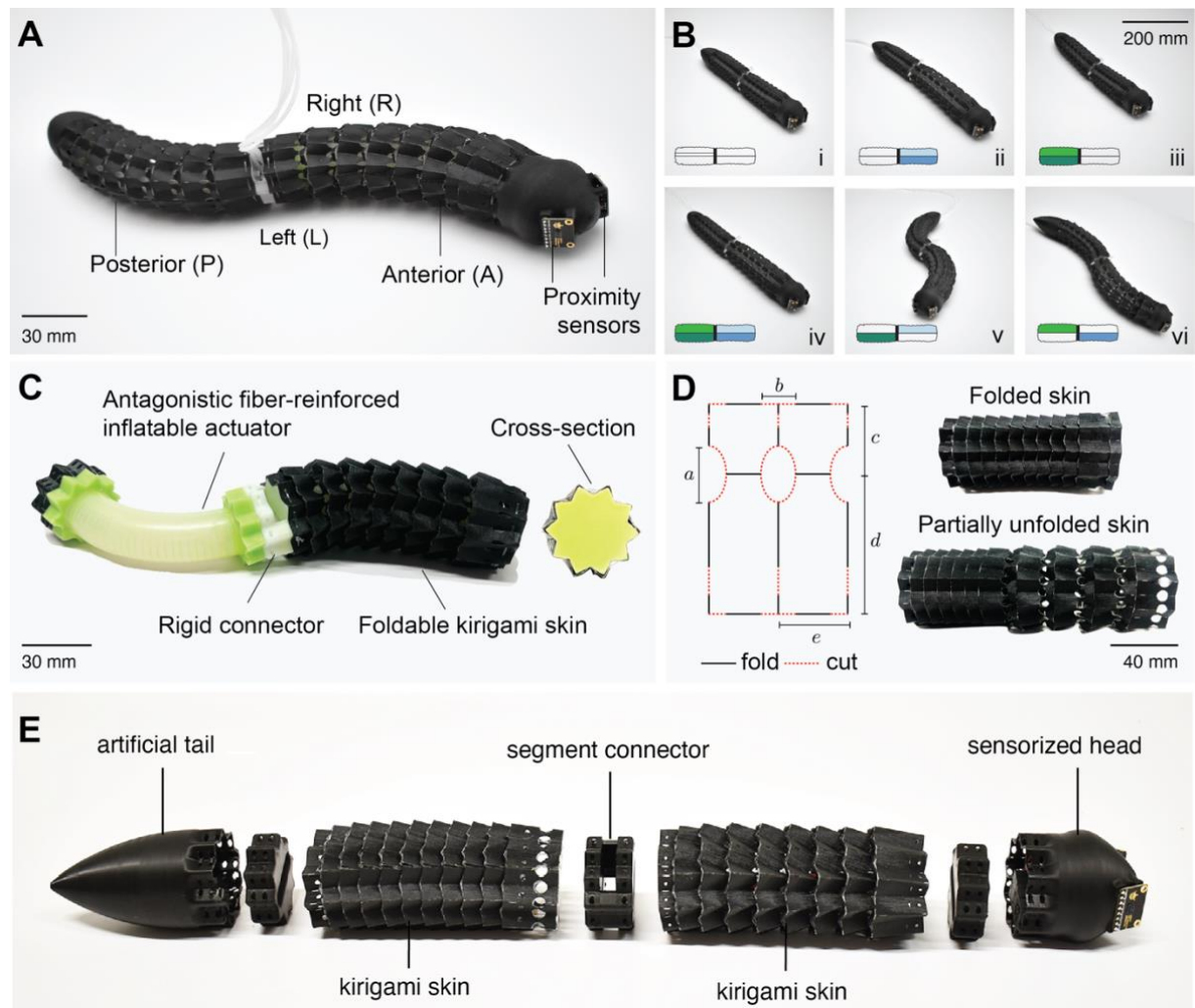

**Fig. 1**. Multimodal crawling robot with kirigami skin. (A) Robot's body featuring anterior and posterior segments and two proximity sensors mounted on its head. (B) Demonstration of different deformation modalities and schematic of inflated chambers. (C) The inner structure of the robot's body showing the assembly of the foldable kirigami skin and the antagonistic fiber-reinforced inflatable actuator. (D) Geometry of the skin unit cell with black fold lines and red cut lines, and snapshots of the multistable foldable kirigami skin in folded and partially unfolded configurations. (E) External components of the robot, including 3D-printed connectors between the anterior and posterior segments, the kirigami skin, the tail, and the sensorized head housing two lateral proximity sensors.

In this work, we propose a limbless crawling soft robot comprising a soft, deformable body made of pneumatic artificial muscles and covered with a foldable kirigami skin designed to break friction symmetry while maintaining high conformability to various combinations of longitudinal and transverse deformations (see Movie S1). The robot is driven by an external pressure source with oscillatory pressurization generated by a Central Pattern Generator (CPG), which can produce stable rhythmic patterns with only simple input signals [27]. Equipped with a rounded head bearing two proximity sensors and a pointy tail, the robot can crawl on surfaces with moderate roughness levels and avoid obstacles through assisted teleoperation (see Fig. 1A). This design demonstrates how the synergy between longitudinal and transverse body deformation and friction modulation enables crawling and maneuvering on uniform terrains. This study introduces a kirigami-based robotic skin possessing three key innovations: (i) the multistability of the design enables large elongations (> 50%) with only small variations in the activation force; (ii) the structural design contributes friction anisotropy for improved locomotion, (iii) through gradual unfolding, uniform deformation without distortion or crumpling is obtained.

**Materials and Methods**

**Design of the antagonistic fiber-reinforced inflatable actuators**

Antagonistic actuators are paired actuators that produce opposing forces to enable controlled, muscle-like movement. The robot's movement relies on two pairs of antagonistic, fiber-reinforced inflatable actuators made of silicone rubber, each featuring two axial chambers for linear expansion and bidirectional bending (see Fig. 1C). These actuators are connected via a 3D-printed PLA connector. The fabrication is performed with an SLA 3D-printed mold comprising four parts that shape the internal and external structures and guidelines for fiber reinforcement. Prepolymer silicone (Ecoflex 00-50) is injected and cured at 40°C for 20 minutes in a convection oven (Binder FD56). After curing, the actuator is removed and placed in a secondary mold for fiber winding using Kevlar threads in a double-helix pattern, allowing controlled linear elongation along its axis without any radial expansion when pressurized. The molded structure is then encapsulated in a plexiglass tube, ensuring concentric alignment, and an additional silicone layer is cast on top to secure the fibers. This is followed by another curing cycle and careful demolding. The actuator is subsequently mounted in a third mold to cast stiffer silicone end caps (Double Elite 32) that match the kirigami skin cross-section, facilitating attachment. After curing at room temperature for 20 minutes, silicone tubes to enable air injection are affixed to the actuator with silicone adhesive (Sil-Poxy) to create an airtight seal (see supplementary Fig. S3).

**Foldable Kirigami skin**

The kirigami-inspired skin was designed with folds and cuts to introduce friction asymmetry. Existing stretchable kirigami skins used in previous limbless crawlers [23] suffer from unpredictable crumpling and kinking when the underlying actuator bends. Here, the inclusion of folds enhances the flexural compliance of the skin and allows for uniform bending while preserving the asymmetric configuration of overlapping cut patterns. The asymmetric folds produce greater backward friction compared to the forward direction, which is essential for generating the propulsive force needed to move across different terrains. The repeated pattern of kirigami cuts distributes friction forces spatially across the surface, minimizing slippage and enhances locomotion compared to isotropic skins. The folded kirigami skin geometry features a pattern of adjacent rectangles with circular and partial cuts that form flexible expandable hinges (see Fig. 1D). This pattern, replicated horizontally and vertically, mimics the segmented structure of a worm, supporting longitudinal expansion and multidirectional bending. The skin was fabricated using PET film (Mylar, 0.1 mm thickness) for structural integrity and a high-performance laminated textile (Dyneema, 102g/sqm, 0.13 mm thickness) for durability, cut with a laser cutting machine, and laminated using heat-press techniques. Double-sided adhesive secures the layers, while prefolding and bonding completes the assembly. This customizable design allows fine-tuning of mechanical properties, optimizing the actuator's flexibility and strength for natural, worm-like motion. The foldable kirigami skin is designed with $20 \times 9$ unit cells. Each unicell has a rectangular geometry with an elliptical cut located between the two foldable segments. The dimensions of the unit cells are defined as $a = 6\,mm$, $b = 4\,mm$, $c = 8\,mm$, $d = 16\,mm$, and $e = 8\,mm$ (see supplementary Fig. S4 for fabrication steps of the kirigami skin).

**Control architecture**

The robot's control architecture consists of three main components as shown in Fig. 2A: (i) a human-machine interface (HMI) operated via joystick for assisted teleoperation, receiving user navigation commands and issuing collision alerts; (ii) a control block with a teleoperation module, Central Pattern Generator (CPG), post-processing for high-level motion commands, and a pneumatic system for low-level actuation; and (iii) the robotic platform with proximity sensors, pneumatic valves, and pressure sensors to collect environmental feedback, control actuation, and monitor pressure.

The teleoperation module translates joystick commands into CPG-based control inputs, interpreting directional commands as specific locomotion modes (e.g., rectilinear, left turn, right turn). The sensor system continuously monitors obstacles within 20 cm, ensuring safe navigation and overriding user commands to stop or turn when an obstacle is detected within 5 cm. When the distance exceeds 20 cm, the robot moves straight; in the 5-20 cm range, differential distance ($\delta = d_R - d_L$) from two sensors triggers steering alerts for right ($\delta > 0$) or left ($\delta < 0$) adjustments (see Note S3, Fig. S5 and Fig. S6 for more information).

The neural control system incorporates a CPG-based controller paired with a postprocessing module as shown in Fig. 2B. The CPG produces diverse rhythmic locomotion patterns using two interconnected neurons, $N_1$ and $N_2$, with modulatory input $m$. Each neuron's behavior was modeled by discrete-time dynamics, where the activity $a_i$ at time $t + 1$ depends on synaptic weights $w_{ij}$ and presynaptic outputs $a_i(t+1) = \sum_{j=1}^{2} w_{ij} o_j(t)$, $i \in \{1,2\}$ [28]. The output follows a hyperbolic tangent function $o_i(t) = \tanh a_i(t)$, with static synaptic weights $w_{11} = w_{22} = w_0$ and modulated synaptic weights $w_{12} = w_1 + m$ and $w_{21} = -w_1 - m$ that respond to input $m$. The default synaptic weights are set to $w_0 = 1.3$ and $w_1 = 0.1$ throughout this work. The postprocessing module compares CPG outputs $o_1$ and $o_2$ dividing the output period into four regions noted as I, II, III, and IV [29] for selective activation of specific chambers and generating a wave motion by introducing a phase shift $\varphi = \frac{nT}{4}, n = 0, \dots, 3$ (see Fig. 2C). The resulting signals are assigned to four robot chambers ($C_{AR}, C_{AL}, C_{PR}, C_{PL}$) enabling three crawling modes (see Fig. S8 and Fig. S9 for more information):

- Rectilinear locomotion: $C_{AR}(I, II) = C_{AL}(I, II) = 1$ ($\varphi = 0$) and $C_{PR}(I, II) = C_{PL}(I, II) = 1$ ($\varphi = 0, \frac{T}{4}, \frac{T}{2}, \frac{3T}{4}$)
- Turning right: $C_{AR}(I, II) = C_{PL}(I, II) = 1$ ($\varphi = 0$) and $C_{AL}(I) = C_{PR}(I) = 1$ $\left(\varphi = \frac{T}{2}\right)$.
- Turning left: $C_{AL}(I, II) = C_{PR}(I, II) = 1$ ($\varphi = 0$) and $C_{AR}(I) = C_{PL}(I) = 1$ $\left(\varphi = \frac{T}{2}\right)$.

This control system supports smooth gait transitions, with frequency tuning from $f = 0.25\ Hz$ to $f = 1.5\ Hz$ and phase shifts in $T/4$ increments. The processed signals trigger on/off valve commands for the pneumatic subsystem, with the control loop running on an Intel Core i7 computer via a Python script.

The pneumatic control subsystem distributes stable air pressure to the robot's antagonistic soft actuators. Each chamber is independently regulated by a pneumatic solenoid valve (Parker 3-way X-Valve, normally closed) for precise control over inflation-deflation cycles. A silent air compressor (KGK LD-50 L) supplies a constant pressure of $P = 140$ kPa. Power for the valves is provided via a motor driver (Pololu A4988), connected to a microcontroller (ESP32-C6), which also collects data from an integrated silicon pressure sensor (MPX5100, NXP) monitoring each of the four chambers. This pressure data ensures safe inflation levels and real-time monitoring. Silicone tubing (1.5 mm inner diameter) connects the actuator chambers to the pneumatic system via barbed and Luer-lock connectors, ensuring secure and reliable airflow.

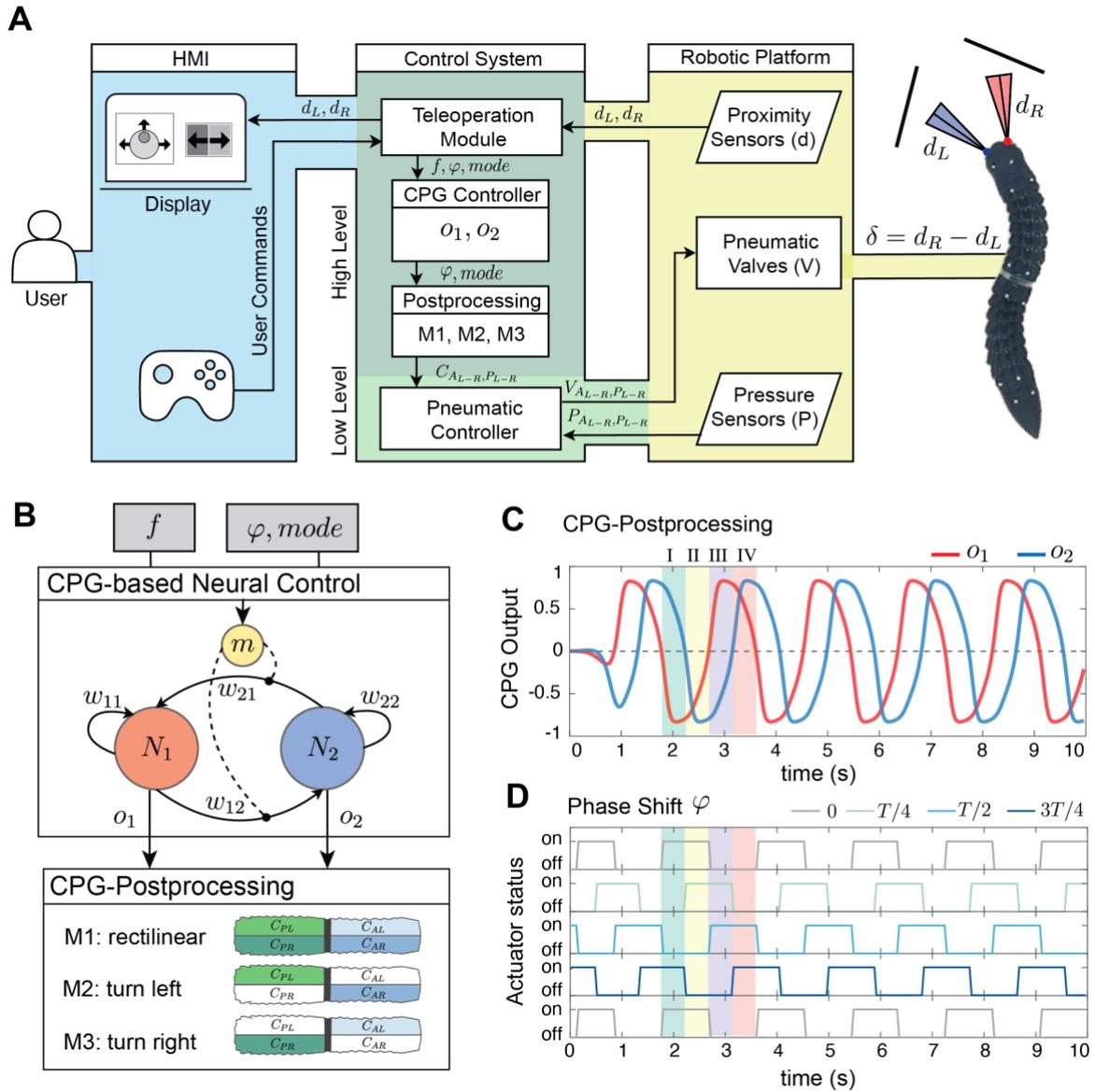

**Fig. 2**. Control architecture. (A) Elements of human-machine interface (HMI), control system, and robotic platform. (B) Structure of the CPG-based controller and postprocessing module. (C) extraction of discrete regions from CPG output. (D) Introducing phase shift to the postprocessed signal for controlling pneumatic valves.

The robot features two distance sensors (Fermion TMF8701 ToF, range: 10-600 mm) mounted in a custom 3D-printed PLA head. These sensors, with a 20° field of view, detect obstacles and aid in path navigation. Positioned with an angular offset of 60° from their central detection axes, they provide a lateral detection field for improved environmental awareness. Real-time sensor data is transmitted via I2C to an ESP32-C3 microcontroller, which compiles and sends the data over WiFi to the high-level control loop, facilitating navigation and teleoperation support (see Note S4, Fig. S7 for more information).

**Results**

**Mechanical response of the kirigami skin**

To characterize the force profile of the kirigami skin during robot actuation, we conducted tensile tests using a universal testing machine (Shimadzu EZ Test). We evaluated kirigami skins with eight rings in two configurations: a single-layer Mylar film and a bilayer of Mylar film with Dyneema fabric. Each prototype had an initial folded length of 110 mm and was stretched between two clamps at a 0.5 mm/s displacement rate. As depicted in Fig. 3A, both prototypes displayed a similar force-displacement response typical of snapping mechanical metamaterials [30]. The force initially increased linearly, followed by sequential snapping as each fold opened, fluctuating between 4.4–6.2 N for the single-layer Mylar and 2.3–4.5 N for the bilayer. Once all folds were open, the force rose steadily. The higher force for the single-layer Mylar resulted from sharp folds that resisted snapping open. With increased fold angles, the bilayer Mylar-Dyneema enabled smoother transitions between states, reducing actuation force. Thus, the bilayer provided a protective layer and facilitated smoother transitions between contracted and expanded states while the snapping behavior limits the force in a controlled range (see Fig. 3B).

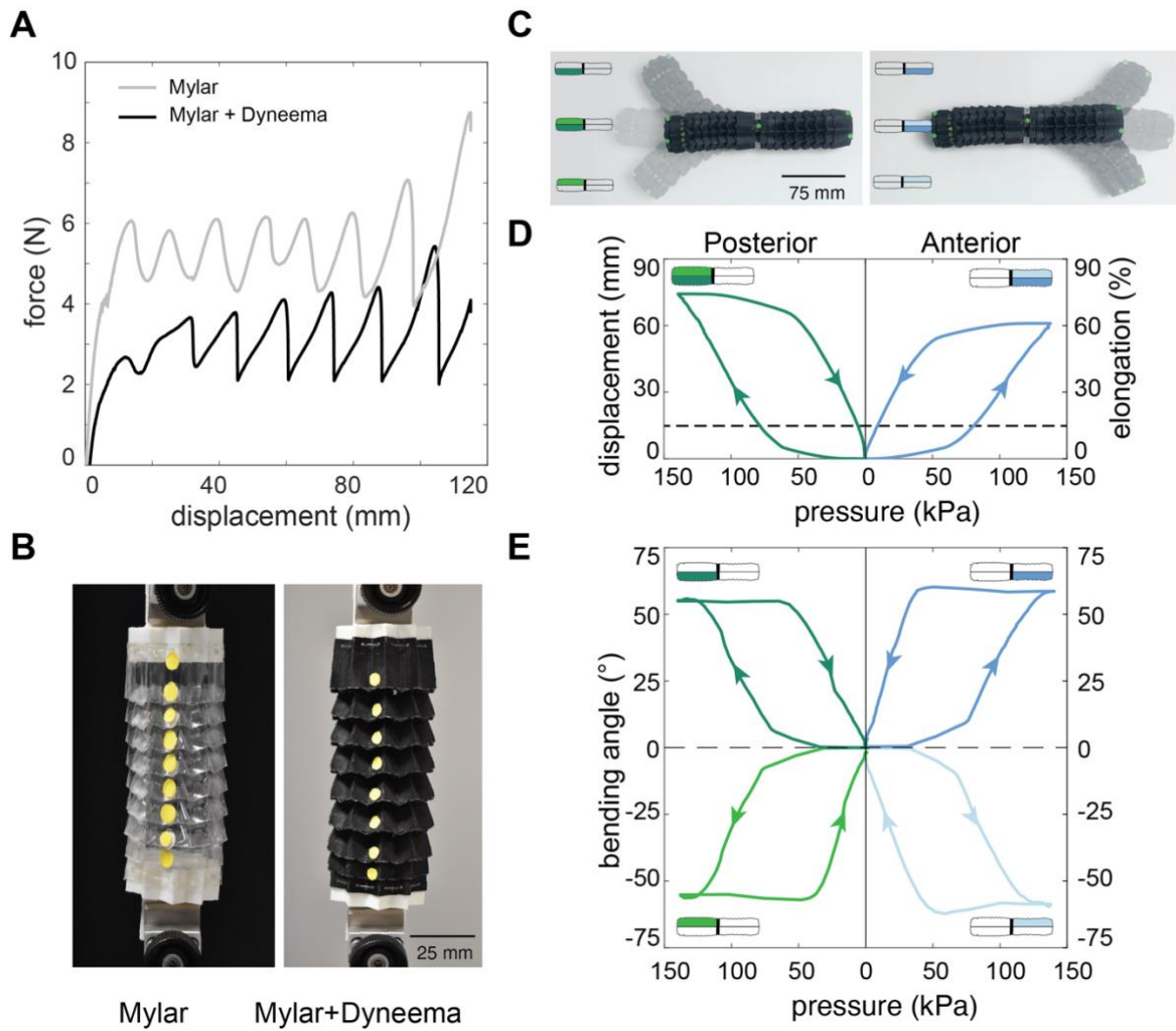

**Fig. 3**. (A) Force-displacement response of the skin for the Mylar layer (gray curve) and a bilayer of Mylar and Dyneema fabric (black curve). (B) Snapshots of the tensile test. (C) Characterization of different actuation modalities. (D) Displacement/elongation, and (E) bending angle profiles of the posterior and anterior actuators.

**Elongation and bending response of the robot**

We characterized the deformation response of the robot to quantify its elongation and bending capabilities (see Fig. Fig. 3C). The robot was fixed at the central rigid joint while its deformation during inflation and exhaustion was recorded using a digital camera. Object tracking software tracked the position of markers attached to the actuator tips. The elongation percentage $\varepsilon = 100 \times (l - l_0)/l_0$ is shown in Fig. Fig. 3D. The posterior segment elongated up to $\varepsilon_P = 74\%$, and the anterior segment reached $\varepsilon_A = 60\%$ at $p_{max} = 140\ kPa$. Each chamber of the anterior and posterior segments was then independently inflated to the same pressure, and the tip angle was measured. Similar behavior was observed in four directions, as shown in Fig. 3E. Maximum bending angles were $\theta_{AR} = 55°$, and $\theta_{AL} = 63°$, $\theta_{PR} = 57°$, $\theta_{PL} = 49°$. Variations across chambers for both elongation and bending were attributed to fabrication inaccuracies. In both experiments, most deformation during inflation occurred after the pressure surpassed a threshold of $p_i \approx 60\ kPa$. During deflation, the actuators began to return to their undeformed state when the pressure dropped below $p_d \approx 50\ kPa$. This behavior is attributed to the minimum pressure required to deform the actuator and the inherent hysteresis in the force-displacement response of the multistable kirigami skin during the opening and contraction of folds.

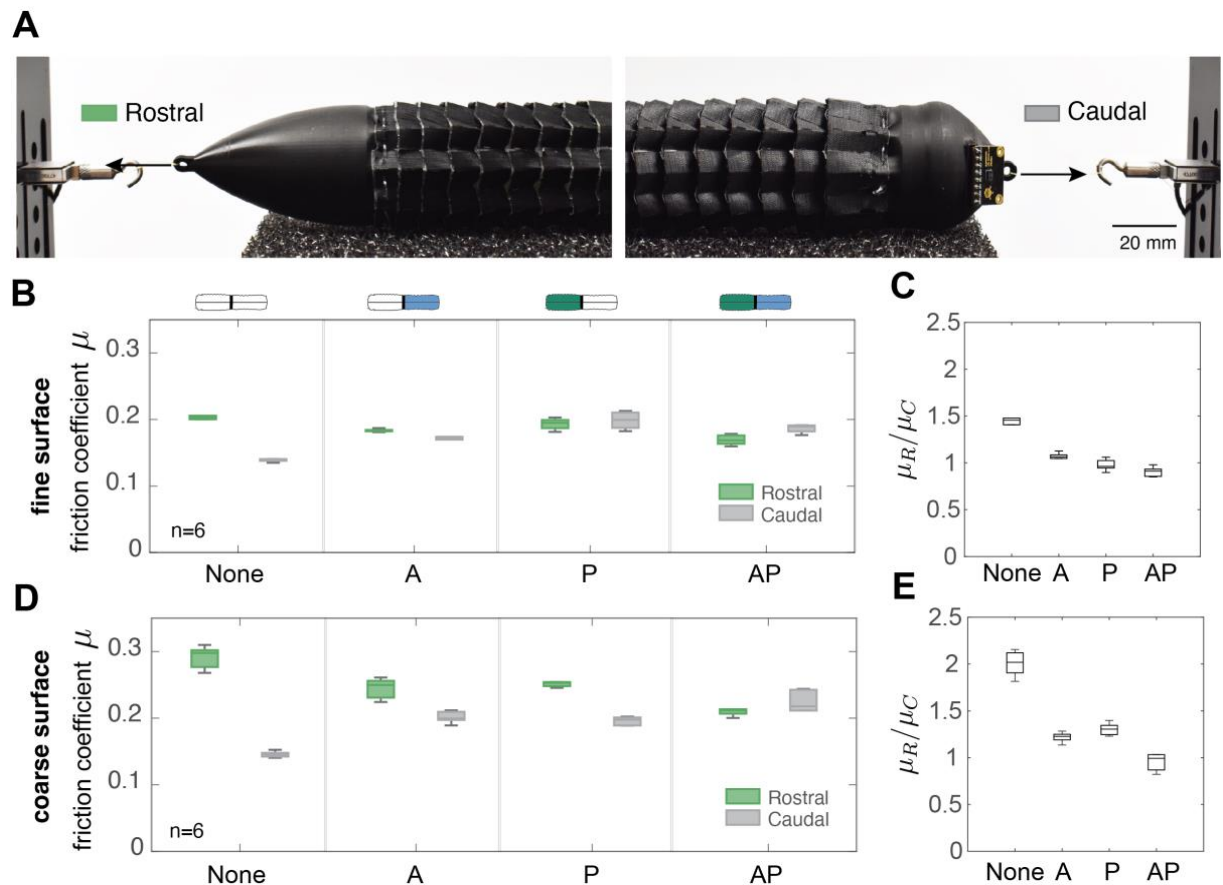

**Fig. 4**. Friction response of the robot. (A) Friction measurement setup for pulling the robot in rostral and caudal directions. (B) Friction coefficients and (C) friction asymmetry ratio of the robot on fine surface (PPI30). (D) Friction coefficients and (E) friction asymmetry ratio of the robot on the coarse surface (PPI10).

**Friction response**

We measured the robot's resistive force when pulled by a Kevlar thread (40 Tex, Aramid) using a motorized linear stage (LTS300C, Thorlabs) against polyurethane foam surfaces with finer (PPI30) and coarser (PPI10) pores with average pore diameters of ø$_{PPI30}$=0.85 mm and ø$_{PPI10}$=2.54 mm, respectively. Friction force was recorded with a load cell (LSB200 S-Beam 5 lbs, FUTEK) mounted on the linear stage as the robot was pulled 200 mm at a constant speed of 10 mm/s. Tests were performed (n=6) for uninflated and three inflation states when anterior (A), posterior (P), and both (AP) segments were inflated, with measurements taken in caudal (along scales) and rostral (against scales) directions (see Fig. 4A). Friction coefficients were calculated as $\mu_C = F_C/W$ and $\mu_R = F_R/W$ where $W = 1.4\ N$ was the robot's weight. Inflation of the robot's segments affected friction force compared to uninflated state (see Fig. 4B and 4D).

Overall, we observed increases in the caudal direction and decreases in the rostral direction, however, for most cases $\mu_R/\mu_C > 1$ which ensures the robot does not slide backward when applying propulsive forces (Fig. 4C and 4E). For some cases, particularly for AP configuration, the friction asymmetry vanishes, i.e., $\mu_R/\mu_C \approx 1$, potentially impacting locomotion. However, these quasi-static tests may differ from dynamic conditions seen during actual movement, highlighting the need for further assessments involving active actuation to better capture performance under real motion scenarios.

**Tuning actuation frequency by adjusting the CPG modulatory input**

The locomotion speed of the robot is intrinsically linked to the inflation and deflation cycles of the antagonistic fiber-reinforced actuators, which follow the oscillation frequency set by the CPG. We mapped the CPG modulatory input parameter $m$ to the desired oscillation frequencies using the pneumatic low-level control system. During this process, we varied $m$ values while monitoring the oscillatory pressure signal with a pressure sensor, from which we extracted the frequency by averaging the period between ten successive peaks. We identified $f_{min} = 0.1\ Hz$ as the minimum frequency that permits a maximum tolerable pressure of $p_{max} = 140\ kPa$. We also determined $f_{max} = 1.5\ Hz$ as the maximum frequency at which the minimum elongation $\varepsilon = 15\%$ required to initiate the opening of the folds of the kirigami skin based on the results reported in Fig. 3D. Following this tuning process, we identified different $m$ values corresponding to selected frequencies in the range $f \in [0.1, 1.5]\ Hz$.

**Rectilinear locomotion**

We measured the speed of the robot while crawling in a straight line on both fine and coarse surfaces for a duration of 1 min (see Movie S3). We conducted these rectilinear locomotion tests (n=5) at different phase shifts $\varphi$ between the actuation of the anterior and posterior segments. The results are summarized in Fig. 5A and 5B for fine and coarse surfaces, respectively. Overall, the robot achieved faster propulsion on the coarse surface than on the fine surface, due to the improved grip provided by the larger pores. The best crawling performance on both surfaces was achieved for $\varphi = T/4$ within the $f = 0.5 - 1\ Hz$ frequency range ($\bar{v}_{max}^{fine} = 6.33\ mm/s$, $\bar{v}_{max}^{coarse} = 10.83\ mm/s$). This optimal range reflects a trade-off between body elongation and oscillatory frequency: lower frequencies enable larger displacements per cycle but fewer cycles, whereas higher frequency yield more cycles with limited per-cycle displacement. Comparing the performance of $\varphi = T/4$ and $\varphi = 3T/4$ reveals that the activation order of body segments plays a critical role in locomotion, despite their similar temporal phase shifts. In the $\varphi = T/4$ protocol, the anterior segment activates before the posterior, maintaining frictional engagement in the posterior segment, which supports propulsion by anchoring the body against the substrate. Conversely, in the $\varphi = 3T/4$ protocol, the posterior segment activates first, followed by the anterior segment, reducing this anchoring effect and diminishing propulsion efficiency.

We developed a simple theoretical dynamic model to qualitatively verify whether the phase shift can influence locomotion speed. The model consists of three interconnected masses linked by two oscillating elements, each exhibiting variable asymmetric friction with the ground (refer to Note S1 for detailed information). Our numerical results demonstrate that the phase shift plays a critical role in determining the distinct dynamic behaviors across the robot body (see Fig. S1). These behaviors are comparable to the tracked position history of three points located at the front, middle, and end of the robot (see Fig. S2). The phase shift significantly alters the friction forces acting on the system, thereby leading to variations in crawling velocities. Specifically, the phase shift modulates the timing and coordination of the oscillating links, which in turn affects how the friction forces are distributed and utilized for propulsion (see Movie S2).

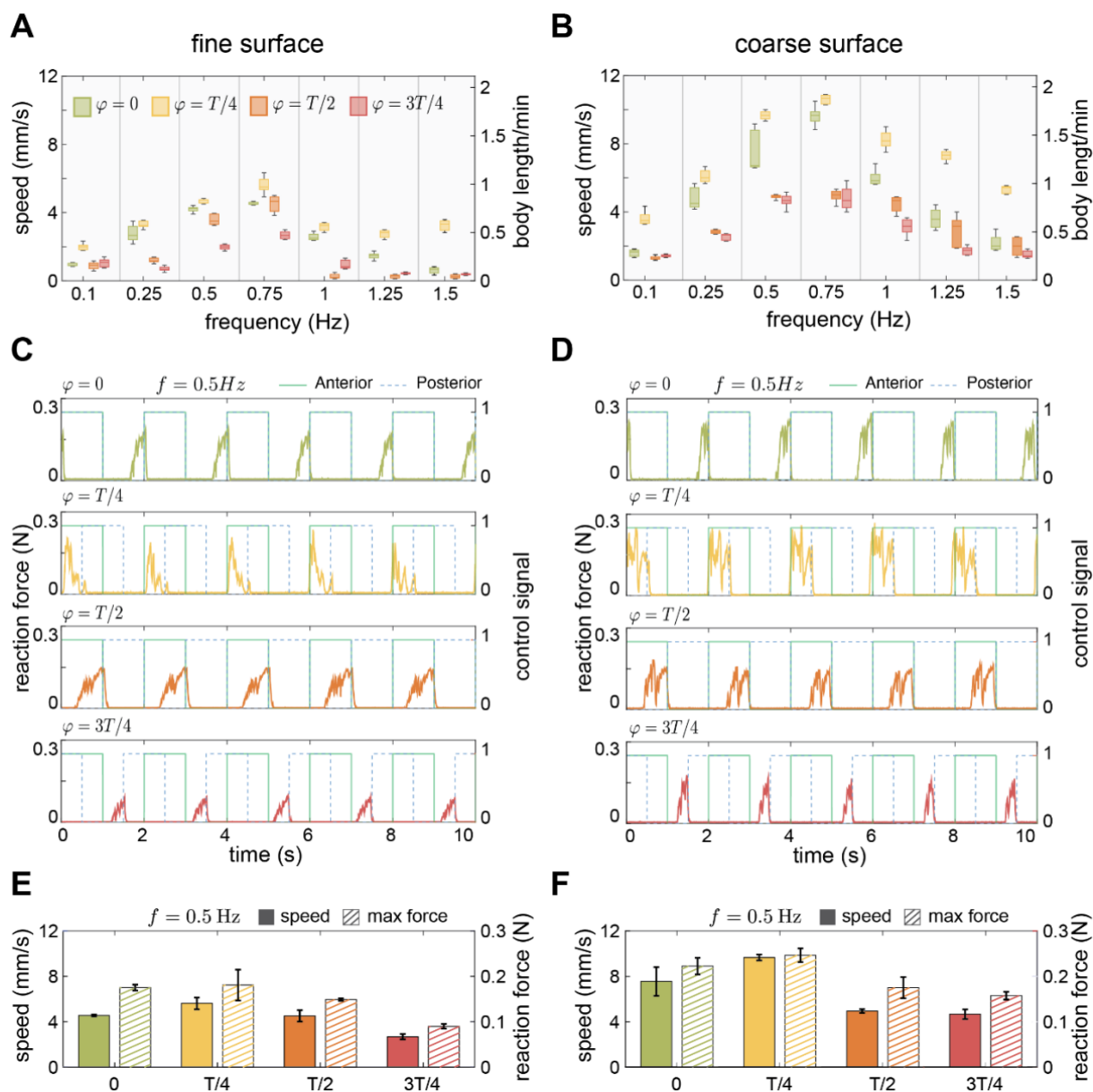

**Fig. 5**. Locomotion speed during rectilinear for different actuation frequencies crawling on (A) fine and (B) coarse surfaces. Pulling force measurements at $f = 0.5$ Hz on (C) fine and (D) coarse surfaces. Correlation of peak forces and corresponding velocity during crawling at $f = 0.5$ Hz on (E) fine and (F) coarse surfaces.

**Dynamic pulling force**

To gain deeper insight into the linear speed results presented in the previous section, we measured the pulling force exerted by the robot under different phase shifts, $\varphi$, while crawling on both fine and coarse surfaces (see Movie S4). All tests were conducted at a frequency of $f = 0.5\ Hz$. For measuring the robot's pulling force, we attached it to a load cell (LSB200 miniature high-performance S-Beam, 5 lbs., FUTEK) using a Kevlar thread at its posterior end, ensuring the thread was relaxed with no preload (see supplementary Fig. S10). The robot was then actuated with four phase shifts ($\varphi = 0, T/4, T/2, 3T/4$) for 2 min. After a few initial cycles, the robot began generating tensile force as it started moving and producing pulling force. This experiment revealed observable differences in both the magnitude and profile of the pulling force across different phase shifts (see Fig. 5C and 5D). We further analyzed the correlation between the measured reaction force and the robot's speed. Fig. 5E and 5F compares the robot's speed and maximum force averaged over 10 dominant peaks on both fine and coarse surfaces across four phase shifts $\varphi$ (see Fig. S10). The results indicate that pulling forces are consistently higher on the coarse surface than on the fine surface, with the highest forces ($\bar{F}_{max}^{fine} = 0.215\ N, \bar{F}_{max}^{coarse} = 0.262\ N$) in both cases observed at $\varphi = T/4$. Interestingly, we observed a correlation between the pulling force and the robot's speed during rectilinear locomotion on both surfaces ($R_{\bar{v}-\bar{F}}^{fine} = 0.77, R_{\bar{v}-\bar{F}}^{coarse} = 0.86$). The correlation between the maximum measured force and locomotion performance can be attributed to the static friction at the anchoring limit.

**Steering modalities**

To navigate arbitrarily complex paths, the robot is equipped with both rectilinear locomotion and steering capabilities. It changes direction using asymmetric actuation in its anterior and posterior segments. Various actuation combinations can be employed to achieve steering (see Movie S5). For example, cyclic inflation of opposite chambers in the anterior and posterior segments enables on-the-spot rotation at high actuation frequencies, as demonstrated in Fig. 6A. At lower frequencies, similar sequences result in wider turns (see Movie S6). In this gait, activating the right chamber of the anterior segment causes the robot to turn to the right, even if the head is oriented to the left, due to the anterior segment bending to the right. Similarly, activating the left chamber of the anterior segment produces a leftward rotation. In this modality, actuating the opposite chamber in the posterior segment creates a torque that drives the rotation.

Another steering strategy is achieved by actuating three chambers. In Fig. 6B, the synchronized posterior chambers and an anterior chamber with a phase shift of $\varphi = T/4$ trigger sidewinding toward the activated anterior chamber. In Fig. 6C, two synchronized anterior chambers and a posterior chamber with a phase-shift of $\varphi = T/4$ enable curvilinear steering in the direction opposite to the activated posterior chamber.

**Navigating through obstacles**

To further showcase the robot's capability for limbless crawling through obstacles, an experimental arena was designed. The arena featured a coarse substrate composed of polyurethane foam (PPI10) and included three strategically placed obstacles. The robot was tasked with navigating from one side of the arena to the other while avoiding collisions. For this task, rectilinear locomotion with a phase shift of $\varphi = T/4$ and a frequency of $f = 0.5\ Hz$ was used for forward progression, while the steering gaits were employed to guide the robot left or right as needed. During the crawling experiment, feedback from onboard proximity sensors was relayed to the human-machine interface (HMI), enabling real-time adjustments to the control inputs to prevent collisions. The robot successfully completed the locomotion task, steering through the obstacles and reaching its goal within 18 minutes (see Movie S7). The trajectory of the robot's movement, along with feedback from the proximity sensors, is visualized in Fig. 6D.

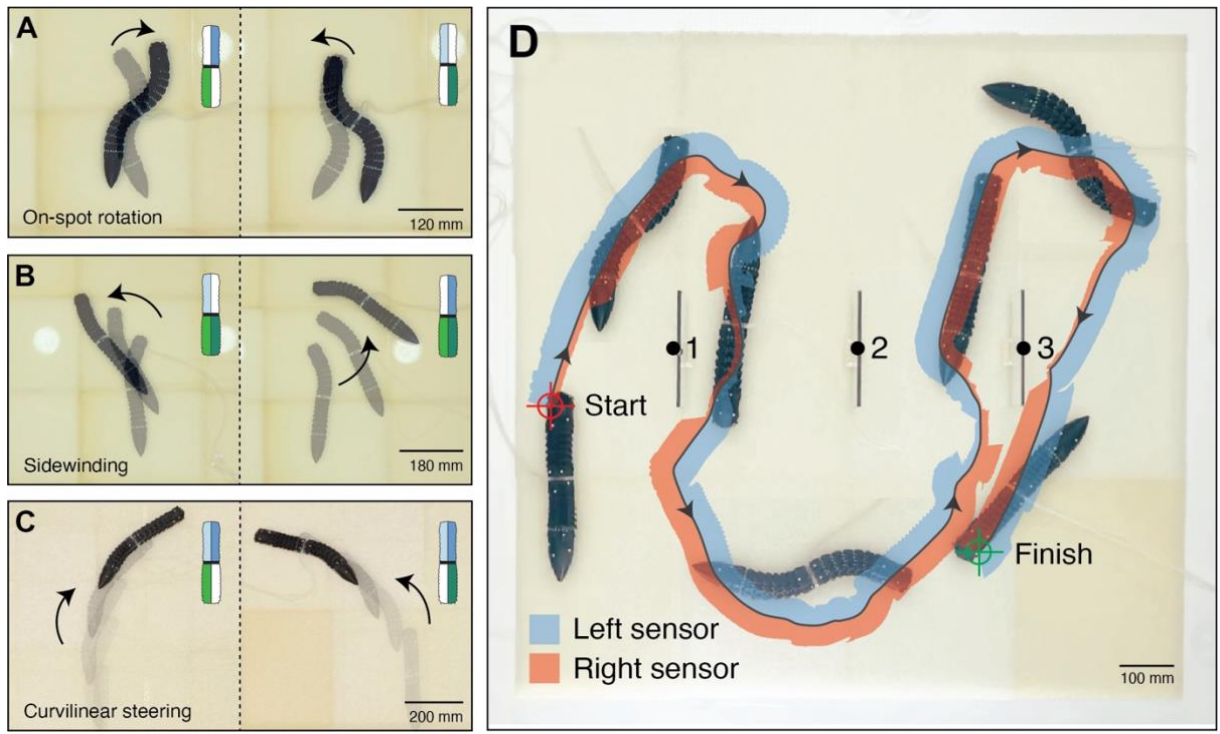

**Fig. 6**. Steering capabilities of the robot. (A) On-the-spot rotation, (B) sideways turning I, (C) sideways turning II, and (D) navigation through obstacles using assisted teleoperation and sensory feedback. The shaded areas indicate distances detected by the proximity sensors, with the values downscaled by a factor of 10 for improved visibility.

**Discussion**

This study demonstrates that combining asymmetric friction with body deformation in a limbless soft robot enables versatile crawling on flat to moderately rough surfaces. The kirigami-inspired skin enhances surface anchoring through directional friction, providing the necessary grip for propulsion. Experimental results reveal that both oscillation frequency and phase control significantly affect locomotion speed, with optimal propulsion achieved when the robot's segments are actuated with a phase offset of one-quarter of the total actuation period. Parametric evaluations identified optimal frequency and phase windows that maximize locomotion performance. A dynamic pulling force response study further established a direct correlation between crawling speed and the force exerted by the robot, with higher speeds under similar control conditions producing stronger pulling forces.

The developed prototype offers a promising foundation for advanced robotic applications, with the flexibility to integrate additional sensors and teleoperation systems. These capabilities enable real-time feedback and obstacle avoidance, making the robot well-suited for navigating complex environments and adapting to external disturbances. The findings provide valuable insights for the design of bioinspired crawling soft robots, particularly for tasks requiring maneuverability in cluttered environments, such as field exploration and search-and-rescue operations.

**Future directions**

To fully realize the potential of combining body deformation with friction modulation in limbless crawling robots, several key areas demand further investigation:

- **Sensor Technology**: Incorporating more advanced and miniaturized sensors to enable finer environmental perception and closed-loop feedback for robust terrain adaptation.
- **Navigation Algorithms**: Developing and testing adaptive control and path-planning algorithms capable of autonomous operation in unstructured and dynamic environments.
- **Material development**: Exploring new materials and structural designs for the kirigami skin to increase durability, tunability of frictional properties, and environmental resilience.
- **Terrain Adaptability**: Conducting experiments across a broader range of surface roughness and inclinations to systematically quantify locomotion performance under extreme and variable terrain conditions.
- **System Integration**: Advancing untethered implementations with integrated power sources and wireless communication to support real-world deployment.

**Conclusion**

In summary, this work introduces a limbless soft robot that leverages kirigami skin and body deformation to achieve efficient and controllable crawling. Through experimental validation, we demonstrate how tuning actuation parameters influences locomotion performance, and we establish a link between speed and generated pulling force. The robot's modular architecture supports sensor integration and teleoperation, positioning it as a candidate for deployment in search-and-rescue and exploratory missions. Future improvements in sensory, control, and material domains will further enhance the robot's functionality and real-world applicability.

**Acknowledgements**

**Funding**: This work was supported by the Villum Foundation through the Villum Young Investigator grant 37499.

**Author contributions**: Conceptualization: J.T., A.P. & A.R., Methodology: J.T., A.P., B.S., D.B., A.R., Investigation: J.T., A.P., B.S., & A.R., Visualization: J.T. & A.R., Funding acquisition: A.R., Supervision: J.J. and A.R., Writing, reviewing, and editing: all authors.

**Competing interests**: The authors declare that they have no competing interests.

**Data Availability**

The MATLAB script for the theoretical model is available as part of the supporting materials at https://github.com/SDUSoftRobotics/2025_CBSYSTEMS_Tirado. The datasets generated and analyzed during the study are available from the corresponding author on reasonable request.

**Supplementary Materials**

Supplementary notes S1-S6
Figs. S1 to S10
Videos S1 to S7

*Supplementary Information*
# Multimodal Limbless Crawling Soft Robot with a Kirigami Skin

Jonathan Tirado[1], Aida Parvaresh[1], Burcu Seyidoğlu[1], Darryl A. Bedford[2], Jonas Jørgensen[1], Ahmad Rafsanjani[1,*]

[1] SDU Soft Robotics, Biorobotics Section, The Maersk McKinney Moller Institute, University of Southern Denmark, Odense 5230, Denmark

[2] Drawstring Origami Ltd., London, UK

* Corresponding author ahra@sdu.dk
Date: May 5, 2025

This document contains:
- Note S1. Theoretical model
- Note S2. Fabrication
- Note S3. Assisted teleoperation
- Note S4. System's architecture
- Note S5. Additional results
- Note S6. Description of supporting videos
- Figures S1-S10

**Note S1. Theoretical model**

We modeled the robot's dynamics with a simple theoretical model consisting of three nodes connected by two oscillating links. The model incorporates directional friction forces, interaction forces between nodes, and gravity to describe the dynamics of the system. Periodic actuation of links generates directional motion. Each node has a mass $m_i$ and moves along a one-dimensional axis. The links between the nodes are modeled as springs with a spring constant $k$. The lengths of the anterior (A) and posterior (P) links elongate periodically with time, simulating actuation:

$$L_A(t) = L_0 + \frac{A}{2}(1 - \cos(2\pi f t)) \tag{S1}$$

$$L_P(t) = L_0 + \frac{A}{2}(1 - \cos(2\pi f t + \varphi)) \tag{S2}$$

Here, $L_0$ is the undeformed length of the links, $A$ is the amplitude of elongation oscillations, $f$ is the actuation frequency, and $\varphi$ is the phase shift between the oscillations of the two links. Each node experiences friction forces that depend on the direction of motion defined by forward friction coefficient $\mu_f$ and backward friction coefficient $\mu_b$. The friction force on each node is given by:

$$F_i^{friction} = -\mu_i m_i g\ \text{sign}(\dot{x}_i), \quad i = A, P, M \tag{S3}$$

where $\mu_i = \mu_f\ \text{if}\ \dot{x}_i > 0$ and $\mu_i = \mu_b\ \text{if}\ \dot{x}_i \leq 0$. Since the friction varies with the reconfiguration of the surface caused by the elongation of each segment, motivated by experiments, we assumed friction coefficients varies linearly with the elongation as

$$\mu_f(L) = \frac{\mu_{f2} - \mu_{f1}}{A}(L - L_0) + \mu_{f1} \tag{S4}$$

$$\mu_b(L) = \frac{\mu_{b2} - \mu_{b1}}{A}(L - L_0) + \mu_{b1} \tag{S5}$$

Where we assumed $L = L_A(t)$ for the anterior node, $L = L_P(t)$ for the posterior node, and $L = (L_A(t) + L_P(t))/2$ for the middle node. The interaction forces between the nodes are modeled as spring forces:

$$F_{PM}^{interaction} = k(x_M - x_P - L_P(t)) \tag{S6}$$

$$F_{MA}^{interaction} = k(x_A - x_M - L_A(t)) \tag{S7}$$

The dynamics of the system are governed by Newton's second law. The equations of motion for the three nodes are:

$$m_P \ddot{x}_P = F_P^{friction} + F_{PM}^{interaction} \tag{S8}$$

$$m_M \ddot{x}_M = F_M^{friction} - F_{PM}^{interaction} + F_{MA}^{interaction} \tag{S9}$$

$$m_A \ddot{x}_A = F_A^{friction} - F_{MA}^{interaction} \tag{S10}$$

The oscillating links ($L_A$ and $L_P$) transfer energy to the nodes, overcoming friction and enabling motion. The difference between forward and backward friction coefficients ($\mu_f$ and $\mu_b$) is crucial for net motion. Higher backward friction ($\mu_b > \mu_f$) ensures that the system moves forward during actuation. The stiffness constant $k$ determines how efficiently the interaction forces propagate through the system. The system exhibits nonlinear behavior due to the sign-dependent friction forces and the time-varying link lengths. We solved these equations numerically using `ode113` in MATLAB, a numerical solver for solving ordinary differential equations (ODEs) with high accuracy with automatic step-size adjustment.

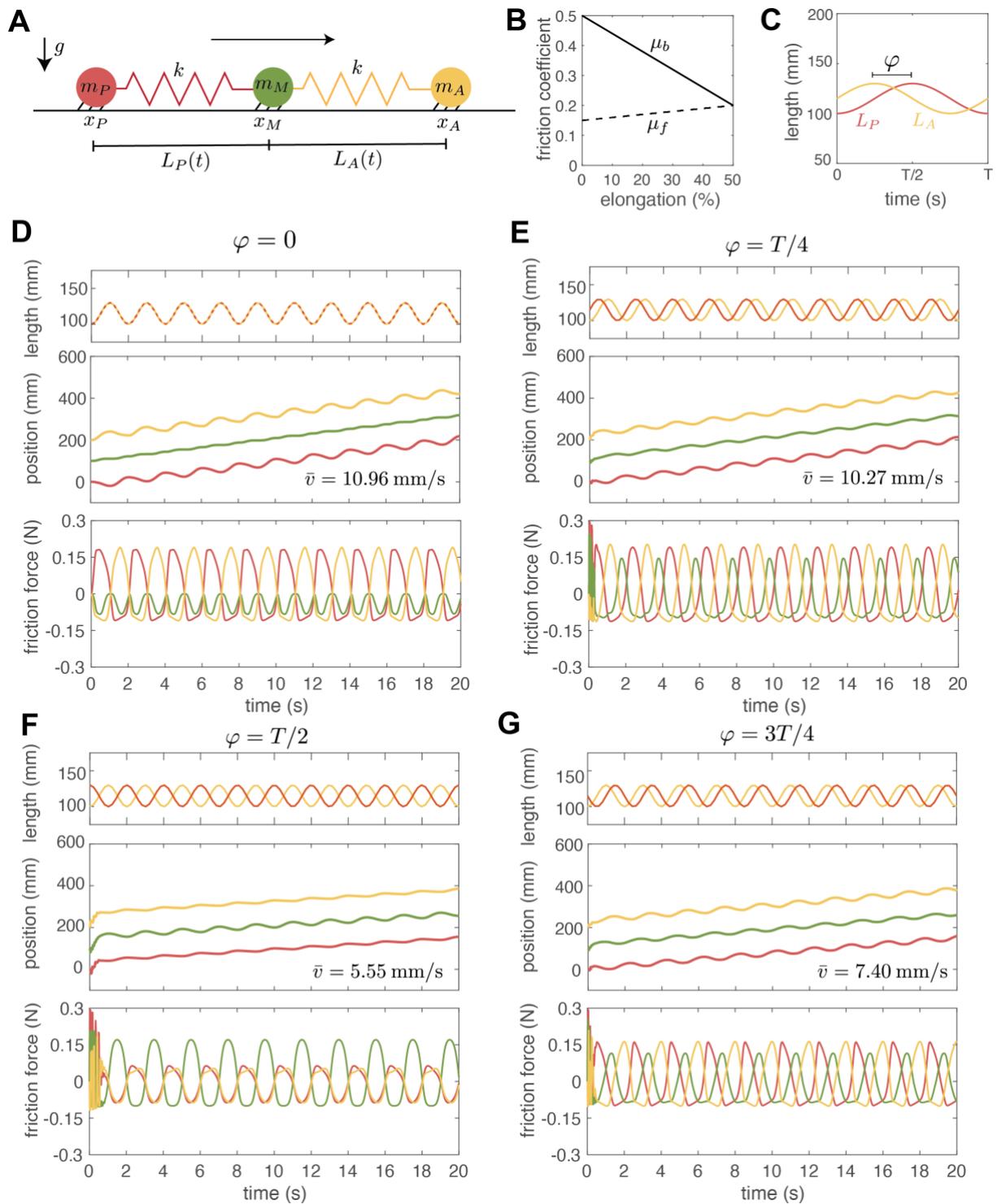

**Fig S1**. (A) Schematic of the theoretical model for a two-segment crawling robot. (B) Friction coefficient as a function of elongation, (C) definition of the phase shift. Imposed actuation length, calculated position of three nodes, and the estimated friction force on each of them are demonstrated for (D) $\varphi = 0$, (E) $\varphi = T/4$, (F) $\varphi = T/2$, and (G) $\varphi = 3T/4$. The color of all plots is in accordance with the schematic in A.

Fig. S1 shows the results for the following parameters: $L_0 = 100$ mm, $A = 0.3\, L_0$, $m_i = 60$ g, $f = 0.5$ Hz, $k = 100$ N/m, $\mu_{f1} = 0.15$, $\mu_{f2} = 0.2$, $\mu_{b1} = 0.5$, $\mu_{b2} = 0.2$, $g = 9.81$ m/s². Here, we assumed a large value for $k$ to ensure the nodes follow the imposed actuation signals $L_A(t)$ and $L_P(t)$ and we increased $\mu_{b1}$ compared to experiments to be able to qualitatively reproduce the observed behaviors in experiments. We also assumed the following approximation: sign $x \approx$ tanh $50x$ to overcome numerical instabilities.

In all models, there is a transient period after which the response reaches a steady state. We ran the simulations for 20 s (10 cycles) and estimated the average speed over the last five cycles. The numerical results shown in Fig. S1 suggest that the robot actuated with $\varphi = 0$ and $\varphi = T/4$ outperformed the cases with $\varphi = T/2$ and $\varphi = 3T/4$. These results are in good qualitative agreement with the experiments, as demonstrated by the crawling of the robot on a coarse surface in Fig. S2. Notably, simply by varying $\varphi$, the model can reproduce comparable velocities and very similar displacement profiles for the three nodes. Additionally, the model enables estimation of the friction forces between each node and the substrate, revealing that varying phase shifts produce distinct friction force profiles (see the third row in Fig. S1D-G).

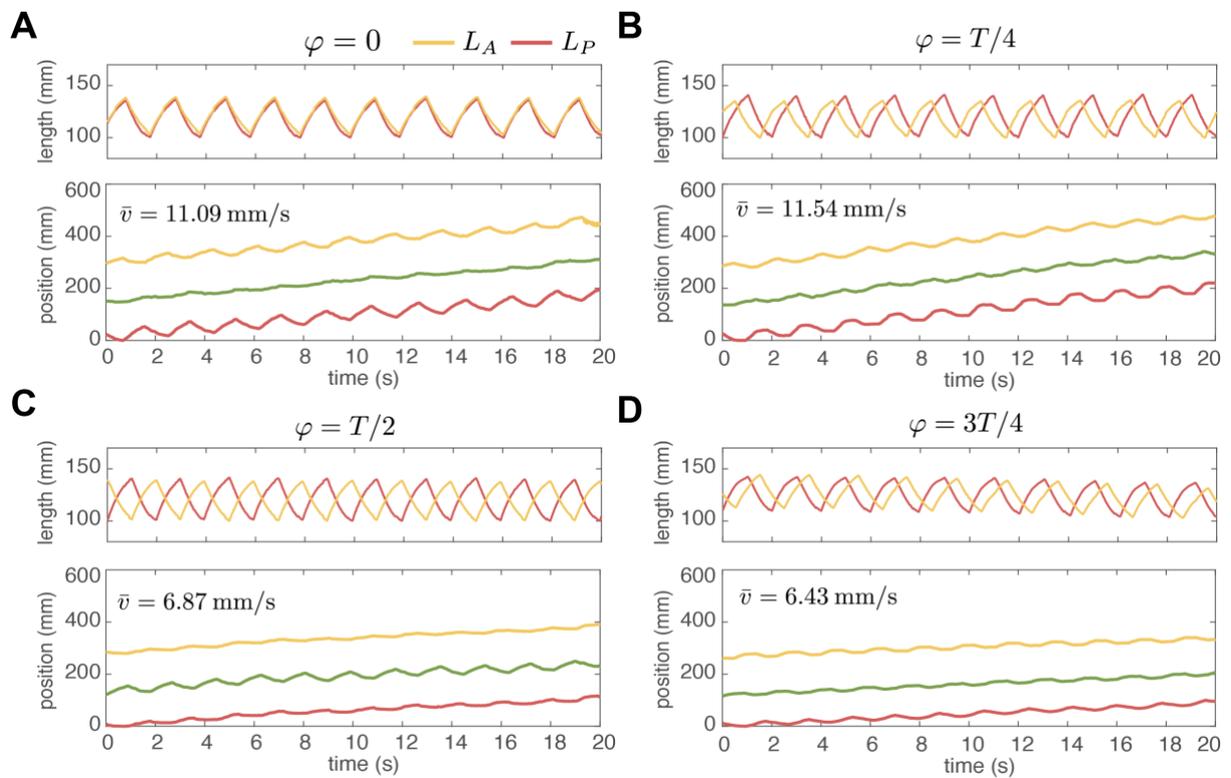

**Fig S2**. Experimental results for crawling of the robot on the coarse surface (PPI10) for different phase shifts. Actuation length for anterior $L_A$ and posterior $L_P$ segment (top) and tracked position of three nodes (bottom) for (A) $\varphi = 0$, (B) $\varphi = T/4$, (C) $\varphi = T/2$, and (D) $\varphi = 3T/4$.

The MATLAB script: https://github.com/SDUSoftRobotics/2025_CBSYSTEMS_Tirado

**Note S2. Fabrication**

The kirigami skin is constructed using a geometric pattern of adjacent unit cells, each composed of rectangular segments with elliptical and partial cuts that form flexible, unfoldable hinges. This pattern is repeated vertically to support longitudinal extension and multidirectional bending. Each unit cell consists of a rectangular geometry, where the elliptical cut, defined by dimensions $a = 6\ mm$, $b = 4\ mm$, is located between two foldable segments. The sides of the unit cell are defined by $c = 8\ mm$, $d = 16\ mm$, and the width is specified by $e = 8\ mm$ (see Fig. S4A). The final structure consists of 20 x 9 unit cells, arranged in two halves (10 x 9 units per half) to maintain the symmetry and avoid undesired bending caused by the stiffened overlapping edges (see Fig. S4B).

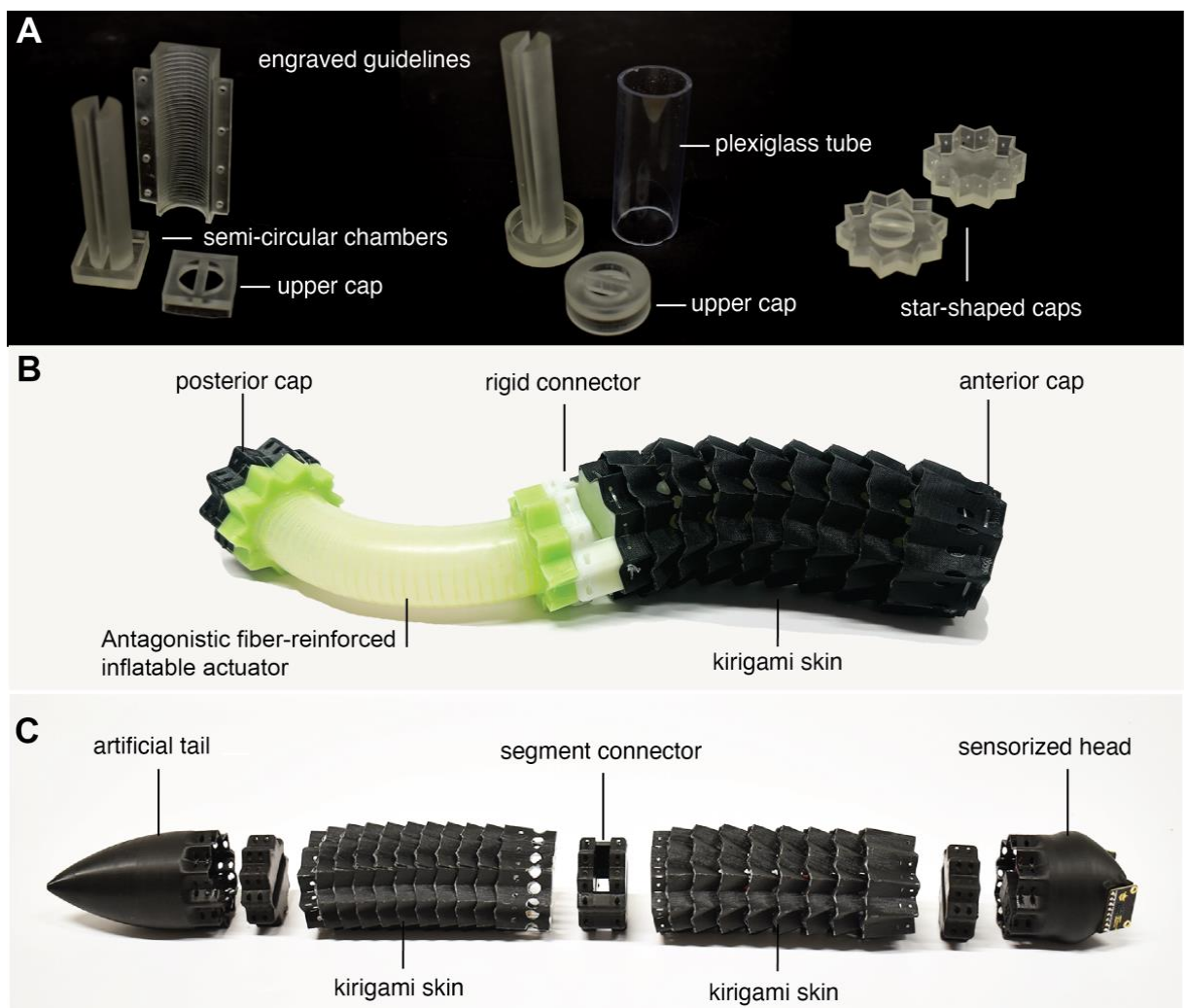

**Fig. S3.** Design and fabrication of the robot. (A) Fabrication of Antagonistic Fiber-Reinforced Inflatable Actuators: The actuators are fabricated using SLA 3D-printed molds and plexiglass tubes to cast prepolymer silicone. (B) Assembly of the Actuator and Kirigami Skin: The robot is constructed with two antagonistic inflatable actuators joined by 3D-printed segment connectors. The kirigami skin is secured to the actuators using anterior and posterior caps that follow the transversal contour of the skin. These caps feature a series of holes, enabling the use of plastic threads to firmly attach the components. (C) External Elements of the Robot: The robot is complemented by a 3D-printed head and tail. Additionally, two proximity sensors are installed on the lateral sides of the head to detect obstacles and enhance navigation capabilities.

The skin is fabricated by combining PET film (Mylar, 0.1 mm thickness) for structural integrity and a high-performance laminated textile (Dyneema, 102 g/m$^2$, 0.13 mm thickness) for durability. The geometric design includes cutting lines (red) and folding lines (black) within the unit cell (see Fig. S4A). These features are created using a laser cutting machine to cut (red) and engrave (black) the PET film, the laminated textile, and a double-sided adhesive that bonds the two structural layers (see Fig. S4D-1).

The assembly process begins by laminating the textile layer with one side of the double-sided adhesive using heat press. After peeling the second adhesive side, the PET film is joined using heat, ensuring a one-column shift between the PET film and the textile layer. This shift creates two overlapping columns for closing the kirigami structure (see Fig. S4D2-4). The two halves are then joined by overlaying one of the free columns (see Fig. S4D-5).

Once the full kirigami pattern is assembled, the vertical guidelines are pre-folded to form a sequence of peaks and valleys, creating a zig-zag configuration (see Fig. S4D6). After folding the complete structure, the second free column is joined, forming a ten-point star prismatic body (see Fig. S4D-7). Finally, the horizontal divisions are folded in a zig-zag pattern, creating the segmented hinges characteristic of the kirigami skin (see Fig. S4D-8). The completed structure enables axial extension and contraction, as well as multidirectional bending (see Fig. S4C).

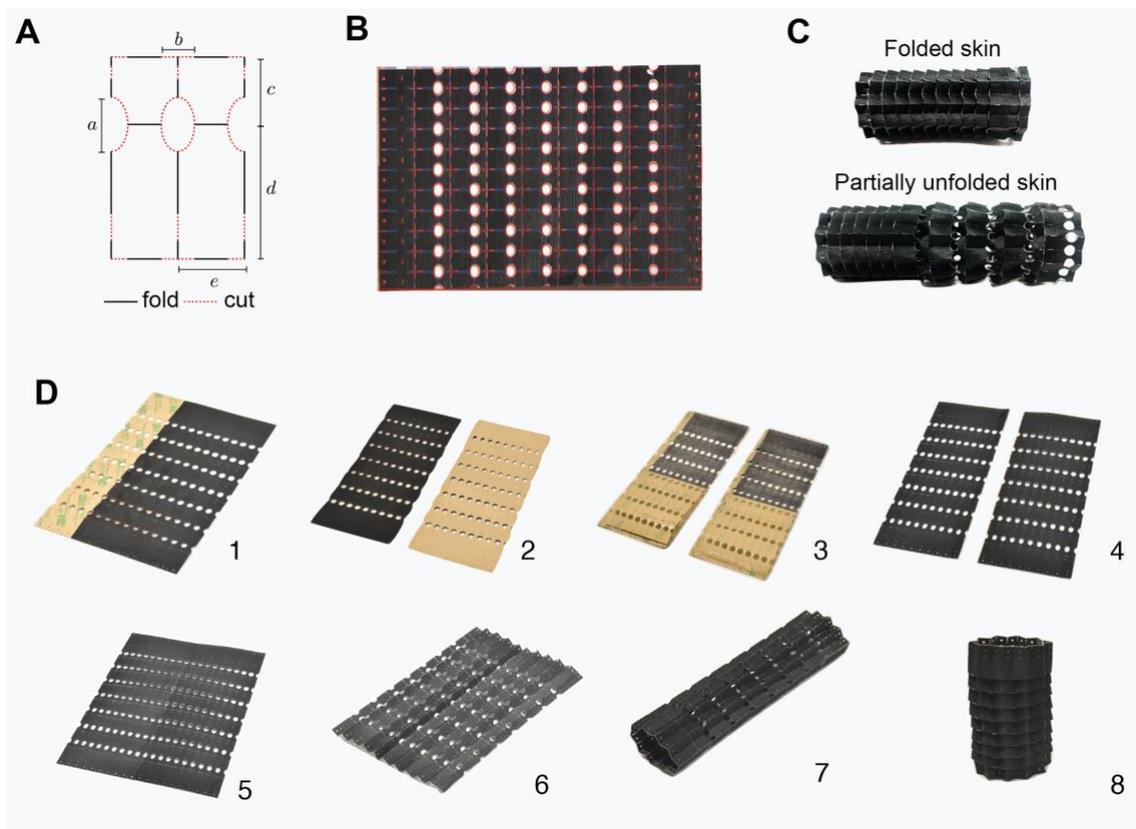

**Fig. S4.** Fabrication of the kirigami skin. (A) Geometry of the unit cell. (B) laser cut skin in flat configuration. (C) Folded skin in closed (top) and partially open (bottom) states. (D) Fabrications steps for the kirigami skin: (1) Mylar film and Dyneema textile layers are cut using a laser cutter, following the kirigami pattern design. (2-4) The Mylar film and Dyneema textile are laminated together using double-sided adhesive and heat press techniques. (5) The upper and lower sides of the kirigami skin are joined using heat-press techniques for seamless integration. (6) The kirigami skin is pre-folded along the vertical guidelines in an alternating sequence, forming the desired 3D structure. (7) An adhesive layer is applied to securely bond the folded sections of the kirigami skin. (8) Finally, the horizontal guidelines are folded to form flexible, expandable hinges of the kirigami skin.

**Note S3. Assisted teleoperation**

The assisted teleoperation platform allows a human operator to navigate the crawler robot while receiving real-time proximity sensor data through a graphical interface on a display shown in Fig. S5.

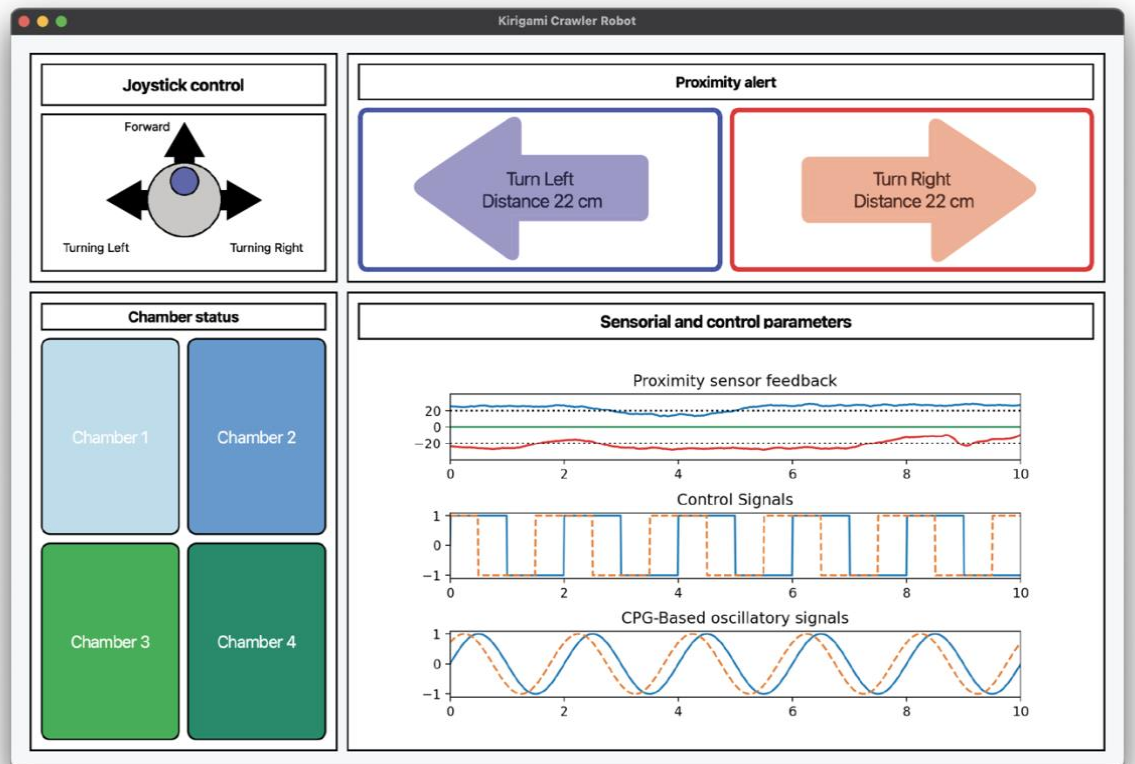

**Fig. S5**. Snapshots of the display used in assisted teleoperation

In addition to user control, the system incorporates a simple obstacle avoidance logic to enhance navigation safety as demonstrated in Fig. S6B.

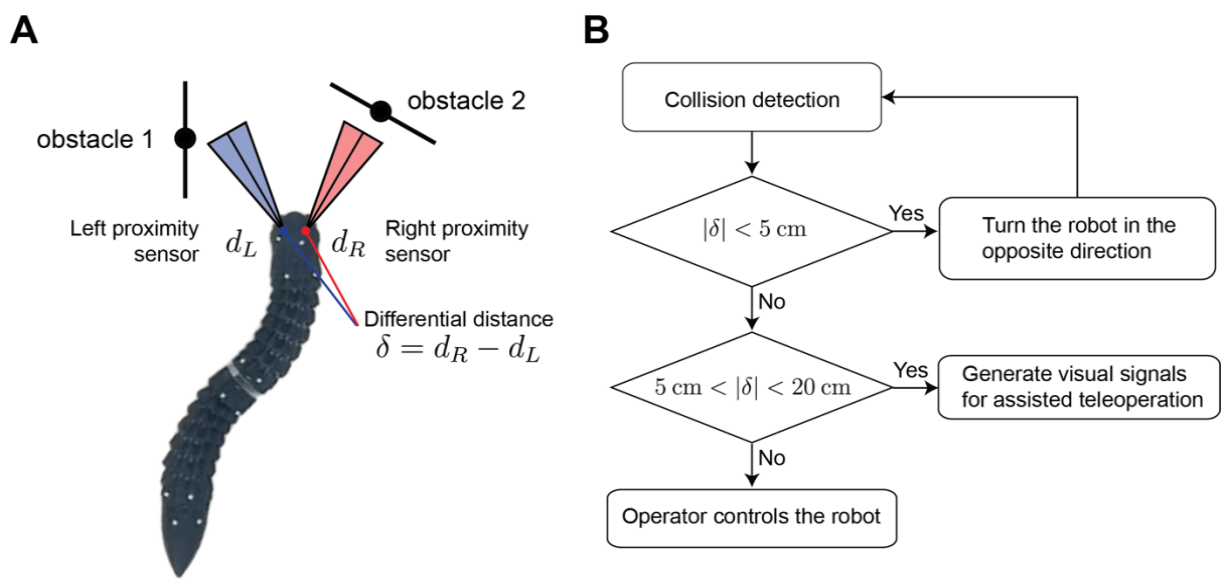

**Fig. S6**. Assisted teleoperation system. (A) Definition of the differential distance from sensor measurements (B) The algorithm for collision detection.

The user interacts remotely with the system via a computer display and a joystick. The display provides visual feedback from the left and right proximity sensors, while the joystick transmits navigation commands to the control system. The teleoperation module processes these inputs and converts them into control parameters, including oscillation frequency, phase-shift, and crawling modality, which serve as inputs for the CPG-based control and post-processing module (see Fig. 2). Two proximity sensors are mounted on the robot's head casing that can detect obstacles within a range of 1 cm to 60 cm. The system operates based on three distance intervals to optimize navigation safety: (i) When no obstacles are detected within 20 cm, the operator has full control over the robot's direction; (ii) If an object is detected within the 5 cm to 20 cm range, the system evaluates the differential distance ($\delta = d_R - d_L$) between the two sensors (see Fig. S5A). A positive $\delta$ prompts the system to alert the operator to steer right, while a negative $\delta$ suggests steering left; (iii) When an obstacle is detected within 5 cm, the system overrides the operator's command and automatically forces the robot to turn in the opposite direction to avoid a collision.

## Note S4. System's architecture

Fig S7 shows the integration of different components of the proposed robotic system including the electronic components, pneumatic system, proximity sensors, and teleoperation system.

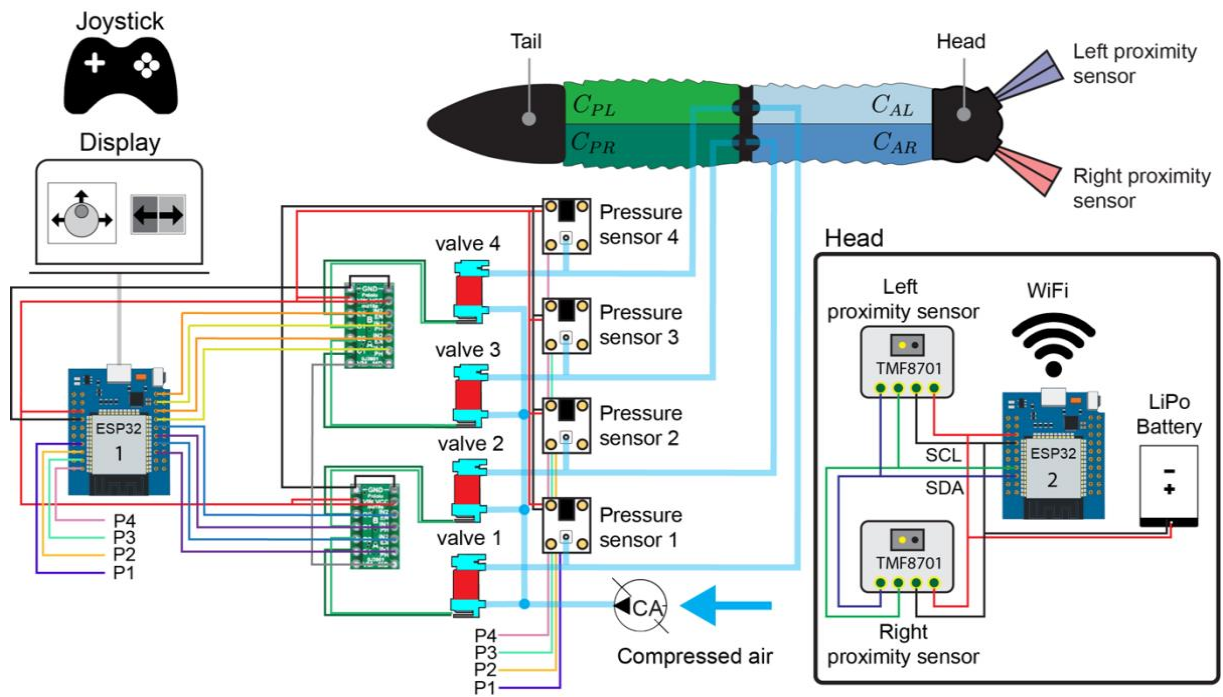

**Fig. S7**. System's architecture. Schematic diagram of the teleoperation system, showcasing hardware components, electrical and pneumatic connections, and wireless communication.

## Note S5. Additional results

In the following we provided additional results on the robot's actuation parameters. Fig. S8 illustrates the output of the central pattern generator (CPG) and the resulting actuation signals for various locomotion modes, including rectilinear movement, sidewinding, and turning. Next, we present the schematic of the robot's actuation sequences for four phase-shifting angles, along with their corresponding pressure responses and variations across different actuation frequencies (see Fig. S9). Additionally, the pulling force exerted by the robot is analyzed over multiple cycles on both coarse and fine surfaces (see Fig. S10).

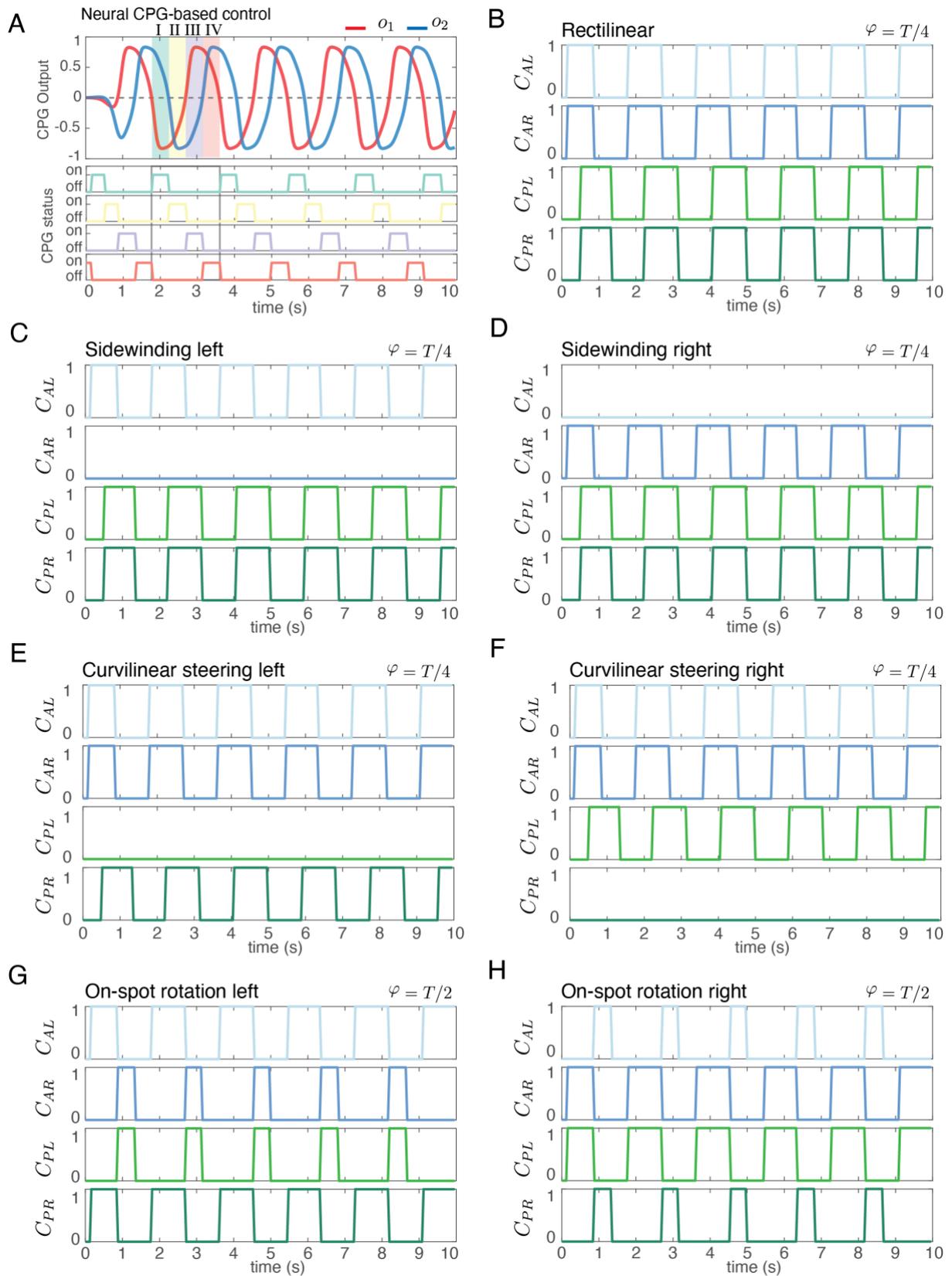

**Fig. S8.** (A) CPG output and postprocessing. Actuation signals for (B) Rectilinear locomotion, (C) winding I left, (D) winding I right, (E) winding II left, (F) winding II right, (G) turning left, and (H) turning right.

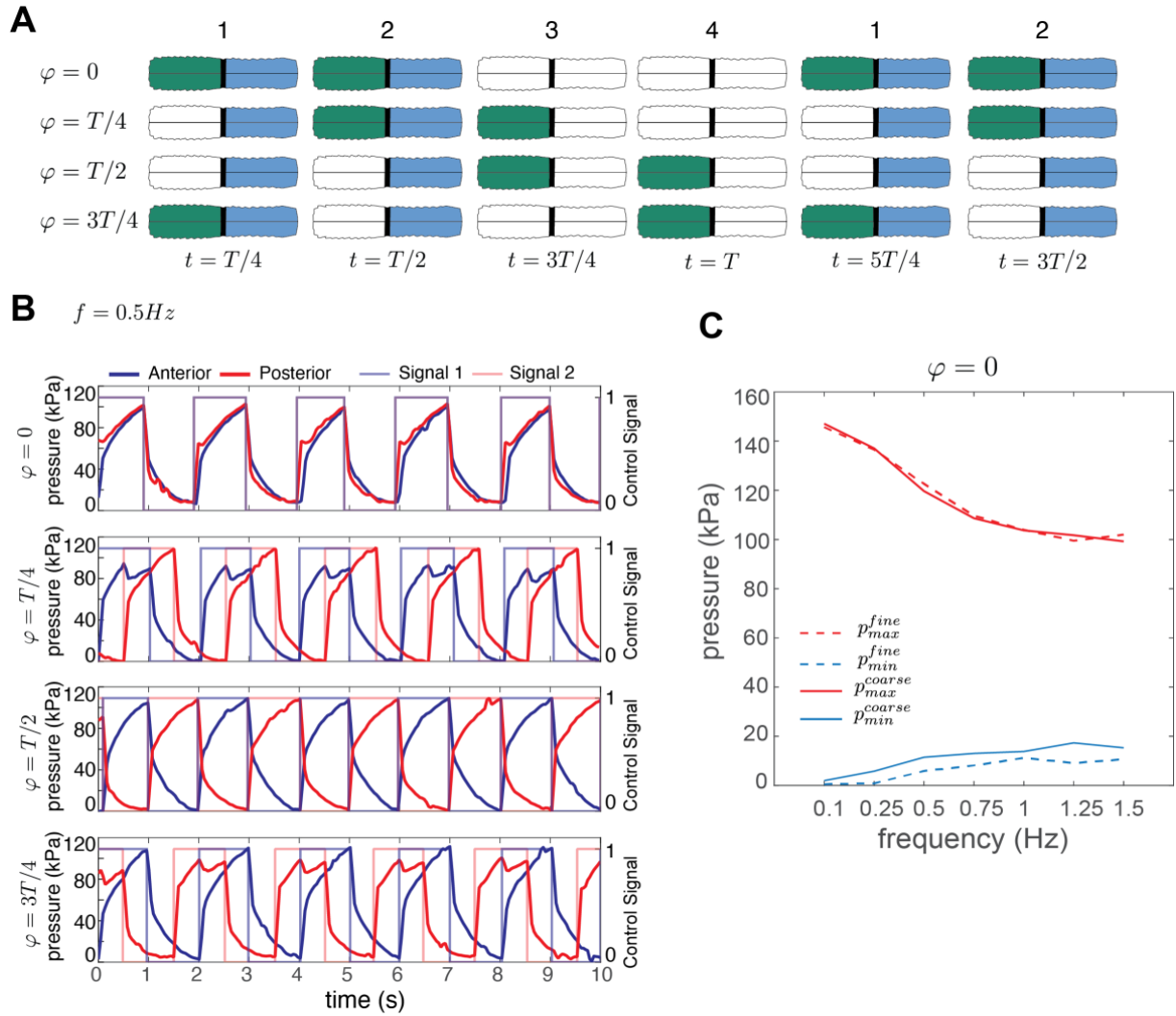

**Fig. S9.** Robot's actuation sequences. (A) Cyclic activation sequence for rectilinear locomotion with different phase shifts $\varphi$. (B) Cyclic pressure profiles for different phase shifts $\varphi$. (C) Minimum and maximum pressures at different actuation frequencies.

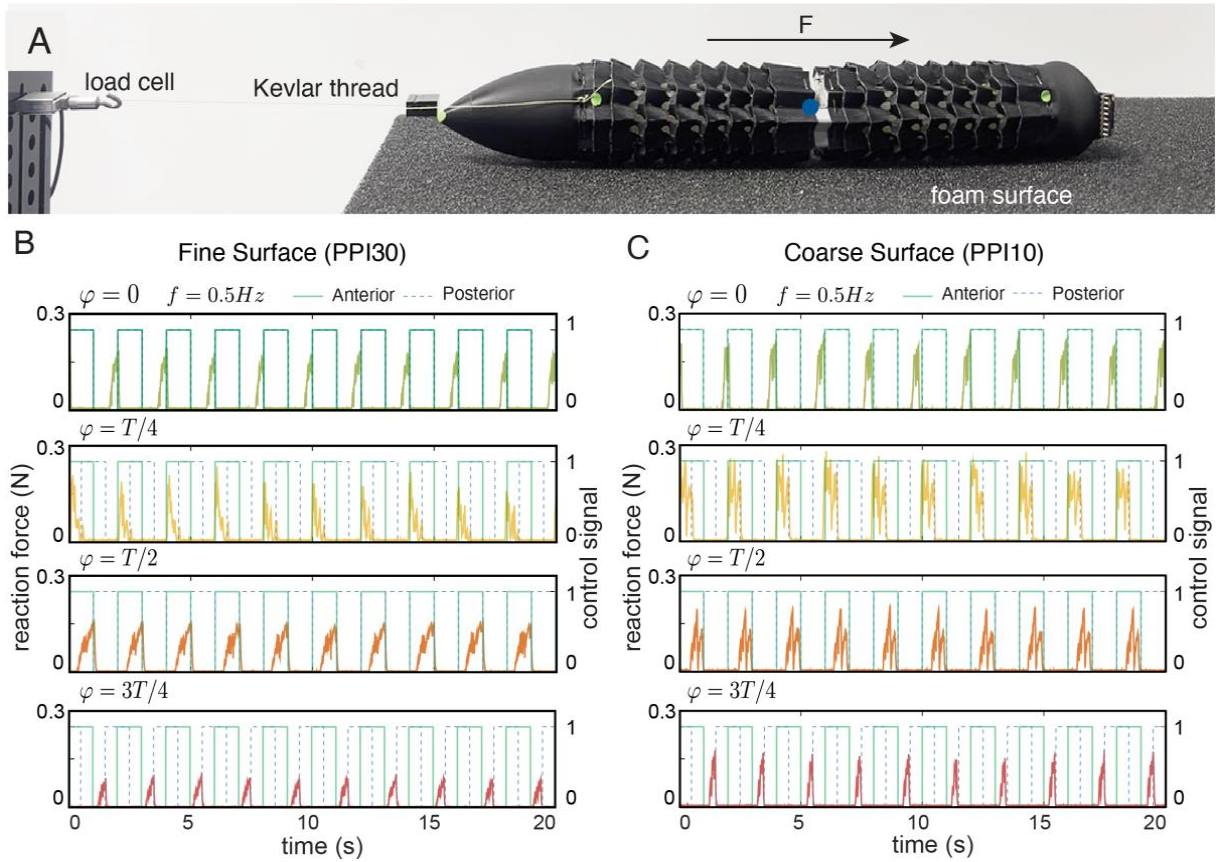

**Fig. S10**. Measured pulling force during locomotion. (A) Experimental setup for measuring pulling force. Pulling force history for multiple cycles on (B) coarse and (C) fine surfaces.

**Note S6. Description of supporting videos**

**Video S1.** Different combinations of actuation sequences of the right and left chambers of the anterior and posterior segments of the robot.

**Video S2:** Rectilinear locomotion. Theoretical model animation

**Video S3**: Rectilinear locomotion. Experimental results

**Video S4**: Traction measurement experiment on coarse and fine surfaces

**Video S5**: Steering modalities for on-spot rotation and sideways turning.

**Video S6**: Complete rotation around a circular path.

**Video S7**: Steering through obstacles with assisted teleoperation and feedback from proximity sensors.